%% file: main.tex
\documentclass[10pt,twocolumn,letterpaper]{article}

\usepackage{cvpr}            
\usepackage{algorithmic}
\usepackage{algorithm}

\usepackage{multirow}
\usepackage{array}
\usepackage{bbding}

\definecolor{cvprblue}{rgb}{0.21,0.49,0.74}
\usepackage[pagebackref,breaklinks,colorlinks,allcolors=cvprblue]{hyperref}

\def\paperID{9275} 
\def\confName{CVPR}
\def\confYear{2025}
\begin{document}
\title{Interleaved-Modal Chain-of-Thought}



\author{%
Jun Gao$^{1}$,Yongqi Li$^{2}$\thanks{Corresponding authors.}, Ziqiang Cao$^{1}$\footnotemark[1], Wenjie Li$^{2}$\\[0.25cm]
{\fontsize{10pt}{12pt}\selectfont $^{1}$School of Computer Science and Technology, Soochow University} \\
{\fontsize{10pt}{12pt}\selectfont $^2$Department of Computer Science, The Hong Kong Polytechnic University} \\
\hypersetup{urlcolor=black}
{\normalsize \href{mailto:jgao1106@stu.suda.edu.cn}{jgao1106@stu.suda.edu.cn}, 
\normalsize \href{mailto:liyongqi0@gmail.com}{liyongqi0@gmail.com}}\\
\hypersetup{urlcolor=black}
{\normalsize\href{mailto:zqcao@suda.edu.cn}{zqcao@suda.edu.cn}, 
\href{mailto:cswjli@comp.polyu.edu.hk}{cswjli@comp.polyu.edu.hk} }\\
{\hypersetup{urlcolor=blue}
\fontsize{9pt}{12pt}\selectfont \href{https://github.com/jungao1106/ICoT}{https://github.com/jungao1106/ICoT}}
}


\maketitle
\input{sec/0_abstract}    
\input{sec/1_intro}
\input{sec/2_related_work}
\input{sec/3_method}
\input{sec/4_experiments}
\input{sec/6_casestudy}
\input{sec/7_conclusion}

{
    \small
    \bibliographystyle{ieeenat_fullname}
    \bibliography{main}
}

\input{sec/X_suppl}

\end{document}

%% file: sec/0_abstract.tex
\begin{abstract}
Chain-of-Thought (CoT) prompting elicits large language models (LLMs) to produce a series of intermediate reasoning steps before arriving at the final answer.
However, when transitioning to vision-language models (VLMs), their text-only rationales struggle to express the fine-grained associations with the original image.
In this paper, we propose an image-incorporated multimodal Chain-of-Thought, named \textbf{Interleaved-modal Chain-of-Thought (ICoT)}, which generates sequential reasoning steps consisting of paired visual and textual rationales to infer the final answer.
Intuitively, the novel ICoT requires VLMs to enable the generation of fine-grained interleaved-modal content, which is hard for current VLMs to fulfill.
Considering that the required visual information is usually part of the input image, we propose \textbf{Attention-driven Selection (ADS)} to realize ICoT over existing VLMs.
ADS intelligently inserts regions of the input image to generate the interleaved-modal reasoning steps with ignorable additional latency.
ADS relies solely on the attention map of VLMs without the need for parameterization, and therefore it is a plug-and-play strategy that can be generalized to a spectrum of VLMs.
We apply ADS to realize ICoT on two popular VLMs of different architectures.
Extensive evaluations of three benchmarks have shown that ICoT prompting achieves substantial performance (up to 14\%) and interpretability improvements compared to existing multimodal CoT prompting methods.

\end{abstract}

%% file: sec/1_intro.tex
\section{Introduction}
\label{sec:intro}
Chain-of-Thought (CoT)~\cite{wei2022chain} prompting aims to augment the reasoning capabilities of large language models (LLMs)~\cite{brown2020language,rae2021scaling,chowdhery2023palm,thoppilan2022lamda} by eliciting them to produce a sequence of intermediate natural language reasoning steps before arriving at the final output.
CoT has proven effective in various reasoning tasks, including arithmetic~\cite{cobbe2021training}, commonsense~\cite{liu2021generated}, and symbolic~\cite{srivastava2023beyond}, and it has become a potential pathway to advanced artificial intelligence, as depicted in GPT-o1~\cite{openai2024o1}.

With the development of vision-language models (VLMs), extending CoT prompting into multimodal CoT to improve the reasoning capabilities of VLMs in vision-related tasks becomes increasingly important~\cite{zhangmultimodal,zheng2023ddcot,wang2024t,mitra2024compositional}.
The initial multimodal CoT attempts~\cite{zhangmultimodal,wang2024t} take as input the fused visual and textual embeddings, and train language models, such as T5~\cite{raffel2020exploring} models, to generate text-only rationales and answers.
In the era of VLMs, introducing triple demonstrations composed of an image with the instruction, textual rationales, and the final output (e.g., answer), has proven effective in sparking the reasoning ability of VLMs~\cite{chen2024m}.
Then, related studies focus on improving the linguistic reasoning ability of VLMs.
Specifically, DDCoT~\cite{zheng2023ddcot} leverages VLMs to deconstruct problems and resolve them respectively, and CCoT~\cite{mitra2024compositional} generates scene graphs to prompt VLMs with object and position description.
SCAFFOLD~\cite{lei2024scaffolding} overlays a coordinate matrix onto the image to prompt the VLMs with relative visual positions.
However, these methods still generate text-only reasoning steps, making it hard to express the fine-grained associations with the origin image exactly.
As shown on the left of Figure~\ref{alg:icot}, textual position descriptions, e.g., at the top, are too rough to identify all fruits (orange and banana).

\begin{figure*}
    \centering
    \includegraphics[width=0.98\linewidth]{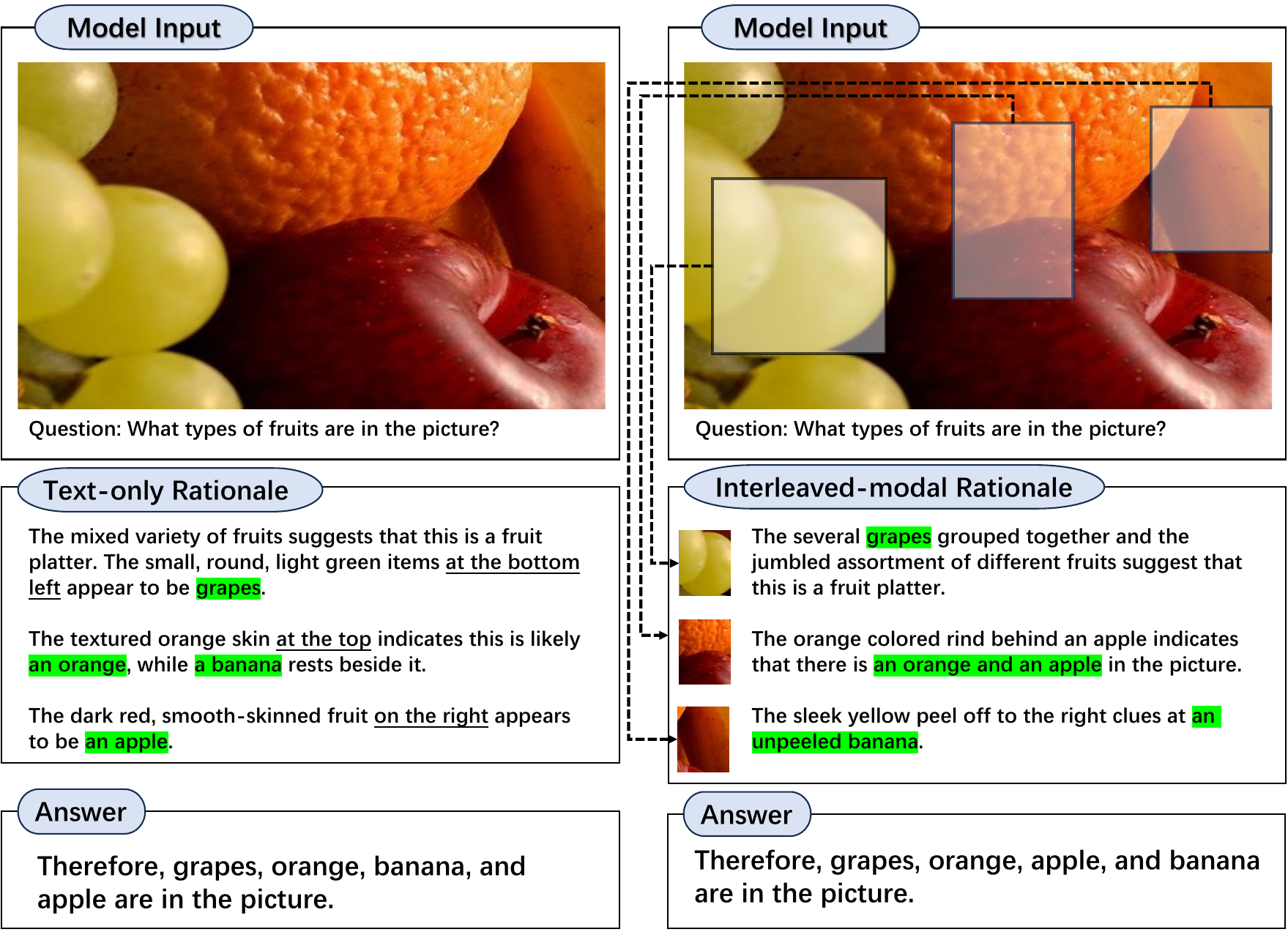}
    \caption{The illustration between multimodal CoT with text-only rationales (Left) and interleaved-modal rationales (Right). \textcolor{green}{Green} blocks are correct texts used to infer the final answer. Text-only rationales restrict VLMs to use a rough description to indicate \underline{the position of objects}. Transparent boxes indicate that these regions are selected and inserted to formulate paired visual and textual rationales in ICoT. }
    \label{fig:icot}
\end{figure*}

In light of the limitations of text-only rationales, we propose incorporating visual information to enhance the precision of fine-grained associations between generated textual rationales and the corresponding image.
We therefore propose an advanced multimodal Chain-of-Thought prompting, named \textbf{Interleaved-modal Chain-of-thought (ICoT)}, as shown on the right of Figure~\ref{fig:icot}.
ICoT generates multimodal rationales consisting of paired images and textual rationales that formulate interleaved-modal reasoning steps to infer the final output.
To the best of our knowledge, ICoT is the first multimodal CoT with images incorporated, and it aligns more closely with human thinking processes~\cite{posner1990attention,bundesen1990theory}.

Intuitively, facilitating the novel ICoT is non-trivial, as it introduces challenges for VLMs to support fine-grained interleaved-modal content generation.
No current VLMs meet this condition completely.
Perceiver-based VLMs such as
Qwen2-VL~\cite{Qwen2VL} converts images into visual embeddings.
Thus, they support fine-grained visual understanding but cannot generate multimodal outputs.
Recently proposed unified-modeling VLMs, such as Chameleon~\cite{meta2024chameleon}, Unified-IO 2~\cite{lu2024unified}, and Emu-3~\cite{wang2024emu3}, enable multimodal generation by tokenizing images into discrete tokens.
However, on the one hand, unified-modeling VLMs exhibit inertia toward multimodal content generation~\cite{chern2024anole}; on the other hand, the generated images belong to the fixed pre-defined resolution instead of fine granularity.

Since required visual information is usually part
of the input image for ICoT, we accordingly propose \textbf{Attention-driven Selection (ADS)} to realize ICoT.
The basic idea of ADS is to signal VLMs to select patches from the input image rather than generating extra images.
At the beginning of generating each textual rationale, ADS inserts a piece of visual tokens of selected patches from the input image to refine the generation of the following textual rationale.
Specifically, ADS utilizes the attention map of VLMs to identify optimal patches from the input image as fine-grained visual rationales.
Once these fine-grained visual rationales are inserted into the current generation sequence, the VLM resumes the original autoregressive text generation process based on previous multimodal content, formulating paired image and textual rationales to infer the final outputs.
Notably, since ADS does not compel VLMs to generate real images, it brings ignorable inference latency compared with previous text-only CoT methods.
Additionally, ADS leverages the attention map of VLMs without requiring parameterization, making it a plug-and-play strategy that can be easily adapted to a wide range of VLMs.

In this paper, we apply \textbf{ADS} to realize \textbf{ICoT} on Chameleon and Qwen2-VL, representing the state-of-the-art unified modeling and perceiver-based VLMs.
The results on existing datasets, including M$^3$CoT~\cite{chen2024m}, ScienceQA~\cite{saikh2022scienceqa}, and LLaVA-W~\cite{liu2024llavanext}, indicate that ICoT realized by ADS brings VLMs with substantial performance gains (up to 14\%) compared with current multimodal CoT methods.
Additionally, it is noted that the tracked interleaved-modal rationales further enhance the interpretability of the generated results.
Our main contribution can be concluded as follows:
\begin{itemize}
    \item We propose interleaved-modal CoT, which innovates text-only rationales into multimodal ones to construct clearer reasoning. To our knowledge, we are the first to incorporate images into the intermediate reasoning steps in multimodal CoT.
    \item We propose an effective and efficient Attention-driven Selection strategy to facilitate ICoT, which is training-free and widely applicable to VLMs without requiring them to support multimodal generation.
    \item Experiments demonstrate that our ICoT significantly surpasses existing multimodal CoT methods, proving that the interleaved-modal reasoning process is a foundational innovation in the line of CoT.
\end{itemize}


%% file: sec/2_related_work.tex
\section{Related Work}

\subsection{Vision-Language Models (VLMs)}
Currently, predominate VLMs such as Qwen-VL~\cite{Qwen2VL,bai2023qwen}, BLIP~\cite{li2023blip}, and LLaVA~\cite{liu2023improved,liu2024visual,liu2024llavanext} are mainly built upon a Large Language Model (LLM), a visual module, and an aligned vision-language adapter.
The visual module, e.g., Vision Transformer (ViT)~\cite{alexey2020image}, encodes images into dense representations, and then the adapter, e.g., MLP or Q-Former, converts these representations into LLM-readable visual tokens.
Finally, visual tokens and textual tokens are fed into the LLM to perform the next-token prediction.
This type of VLM can be concluded as Perceiver-LLM architecture, while the Perceiver usually comprises a visual module and the adapter.
Additionally, Cambrain-1~\cite{tong2024cambrian} introduces more visual modules to collaboratively provide more useful visual tokens in a vision-centric paradigm.
In the other research line, unified-modeling VLMs represented by Chameleon~\cite{meta2024chameleon}, Unified-IO 2~\cite{lu2024unified}, and Emu3~\cite{wang2024emu3} are designed to generate texts, images, and so on uniformly. 
As these models apply codebook~\cite{esser2021taming} to tokenize images into discrete vokens, their training processes are supervised by both vision and text information.
Unified-modeling VLMs are expected to develop more stable multimodal understanding abality~\cite{diao2024unveiling}.


\subsection{Multimodal Chain-of-Thought Prompting}
Similar to CoT used in LLMs, multimodal Chain-of-Thought prompting methods~\cite{zheng2023ddcot,wang2024t,mitra2024compositional,lei2024scaffolding,yang2023set} aim to augment the reasoning ability of VLM by generating intermediate reasoning steps.
A series of studies focus on providing VLMs with fine-grained textual information, such as detailed description~\cite{wang2024t}.
Compositional CoT (CCoT)~\cite{mitra2024compositional} prompts VLMs to generate a Scene Graph (SG), which is a JSON-like description containing compositional information of objects that occurred in the image.
DDCoT~\cite{zheng2023ddcot} deconstruct problems into small problems, requiring VLMs to solve them respectively and then inferring the final answer.
In the other research line, 
Set-of-Marks prompting~\cite{yang2023set} augments the objects in the image to help VLMs recognize them.
SCAFFOLD~\cite{lei2024scaffolding} overlays coordinate onto images to prompt VLMs with relative position information, and VLMs leverage overlayed textual coordinates to implicitly represent corresponding regions of the image to perform reasoning.

However, these methods still produce text-only rationales to infer the final answer.
These generated textual rationales usually struggle to express the fine-grained associations with the origin image.
We thereby propose ICoT to elicit VLMs to generate interleaved visual-textual reasoning steps to effectively reach the final outputs.

%% file: sec/3_method.tex
\section{Methodology}
To address the limitations that current multimodal CoT methods are still stuck in generating text-only rationales to infer the final answer, we propose interleaved-modal CoT (ICoT) to elicit VLMs generated multimodal reasoning steps.
We start by introducing the workflow of VLMs and multimodal CoT in Section~\ref{sec:preliminary}.
We then introduce the concept of ICoT in Section~\ref{sec:icot}.
Finally, we propose a plug-and-play method, Attention-driven Selection (ADS), to realize ICoT on existing VLMs.
\subsection{Preliminaries}
We first recall some background of VLMs and multimodal CoT in this section.

\paragraph{Vision-Language Model.} VLMs usually consist of a visual encoder $\mathbf{E}$ and a generative large language model LLM, and they determine where to insert images according to visual holders inserted in the text instructions.
Then, VLMs take the image and the instructions as input and respond with a final answer
\begin{equation}
  \mathrm{answer} = \mathbf{VLM}(\mathrm{Image}, \mathrm{Instruction}).
  \label{eq:vlm}
\end{equation}
Specifically, the visual encoder $\mathbf{E}$ extracts visual tokens $f_v^{l\times d}$ from the image $x_v$, where $l$ is the length of visual tokens and $d$ is the dimensions of the hidden states of the LLM.
The built-in LLM predicts next-tokens in a left-to-right fashion according to visual tokens $f_v^{l\times d}$ and the instructions.

\paragraph{Multimodal CoT.} 
Compared with the direct prediction described in Eqn.~\ref{eq:vlm}, Multimodal CoT further introduces a prompt to elicit VLMs to generate a series of intermediate textual rationales $(r_1, r_2, ...)$ before the final answer:
\begin{equation}
   r_1, r_2, ..., \mathrm{answer} = \mathbf{VLM}(\mathrm{Prompt}, \mathrm{Image}, \mathrm{Instruction}).
\end{equation}
Technically, the prompt could be represented as a sequence of demonstrations, each consisting of a triple: ($\mathrm{Image, Rationale, Answer}$).
Alternatively, an explicit instruction, such as ``Let's think step by step," could also serve as the prompt.

\label{sec:preliminary}
\subsection{Interleaved-modal Chain-of-Thought}
\label{sec:icot}
\begin{algorithm}[t!]
\caption{Interleaved-modal CoT}
\begin{algorithmic}[1]
    \STATE \textbf{Input:} Word embeddings $f_e$, Visual tokens $f_v$, Selected number $n$, Signal tokens $\mathcal{S}$, Stopping criteria $SC$
    \STATE \textbf{Output:} Generated Response $Answer$
    \STATE $\mathrm{predicted\_tokens} \gets []$ \textcolor{blue}{\COMMENT{Initialize as an empty list}}
    \STATE $\mathcal{V} \gets f_v$

    \STATE $\mathrm{inputs}$ = Initilize($f_e$, $f_v$) \textcolor{blue}{\COMMENT{Initialize inputs for prefilling}}
    \WHILE{$SC$ not met}
        \STATE $\mathrm{next\_token}, \mathrm{attention\_map}$ = \textbf{model}($\mathrm{inputs}$ )
        \STATE Append $\mathrm{next\_token}$ to $\mathrm{predicted\_tokens}$
        
        \textcolor{blue}{\COMMENT{ADS judgement}}
        \IF{$\mathrm{predicted\_tokens} = \mathcal{S}$} 
            \STATE $\mathcal{V}_{\mathrm{selected}}$= ADS($\mathcal{V}$, $\mathrm{attention\_map}$, $n$) \textcolor{blue}{\COMMENT{Apply Attention-driven Selection}}
            \STATE Append $\mathcal{V}_{\mathrm{selected}}$ to $\mathrm{predicted\_tokens}$
            
        \ENDIF 
        \STATE $\mathrm{inputs}$ = Update($\mathrm{inputs}$, $\mathrm{predicted\_tokens}$) \textcolor{blue}{\COMMENT{Updates inputs for next step generation}}
    \ENDWHILE
    \STATE $Answer$ = Tokenizer.decode($\mathrm{predicted\_tokens}$)
    \STATE \textbf{return} $Answer$
\end{algorithmic}
\label{alg:icot}
\end{algorithm}

Previous multimodal CoT prompting methods only produce text-only reasoning steps to improve the reasoning ability of VLMs.
These intermediate steps are generated according to the entire image, which are struggle to express exact fine-grained associations with the original image.
Given these limitations, we propose a more advanced Interleaved-modal Chain-of-Thought (ICoT) prompting, aiming to elicit VLMs to generate a series of multimodal intermediate reasoning steps each consisting of paired image and textual rationale.
Generated intermediate reasoning steps formulating interleaved-modal rationales to effectively lead to the final outputs.
In this paper, we consider the visual rationales in interleaved-modal rationales as fine-grained visual information $x_v^{h' \times w'}$ extracted from an image $x_v^{h\times w}$\footnote{The images in the dataset involved in this paper are RGB images by default, and the number of channels is omitted in the formulas for simplicity.}.
These visual rationales capture relevant details in the image, such as objects, colors, and texts, interleaved with the following generated textual rationale to infer the final answer:
\begin{equation}
\begin{aligned}
    r_1, x_{v_1}, r_2, x_{v_2}, ..., \mathrm{answer} = \mathbf{VLM}(\mathrm{Prompt}, \mathrm{Image}, \\ \mathrm{Instruction}).
\end{aligned}
\end{equation}

\begin{algorithm}[t!]
\caption{Attention-driven Selection}
\begin{algorithmic}[1]
    \STATE \textbf{Input:} Attention map $A_t$, Selected number $n$, Visual tokens $f_v$
    \STATE \textbf{Output:} Fine-grained visual information $\mathcal{V}_{\mathrm{selected}}$

    \STATE $\mathcal{V}_{\mathrm{selected}} \gets \emptyset$ \textcolor{blue}{\COMMENT{Initialize as an empty set}}

    \STATE Indices $ \gets \mathrm{TopK}(A_t, n) $
    
    \FOR{$i$ in Indices}
        \STATE Append $f_v^{l\times d}[i]$ to $ \mathcal{V}_{\mathrm{selected}}$
    \ENDFOR
    
    \STATE Restore($\mathcal{V}_{\mathrm{selected}}$, Indices) \textcolor{blue}{\COMMENT{Restore relative positions in-place}}
    
    \RETURN $\mathcal{V}_{\mathrm{selected}}$
\end{algorithmic}
\label{alg:ads}
\end{algorithm}

\begin{figure}
    \centering
    \includegraphics[width=0.99\linewidth]{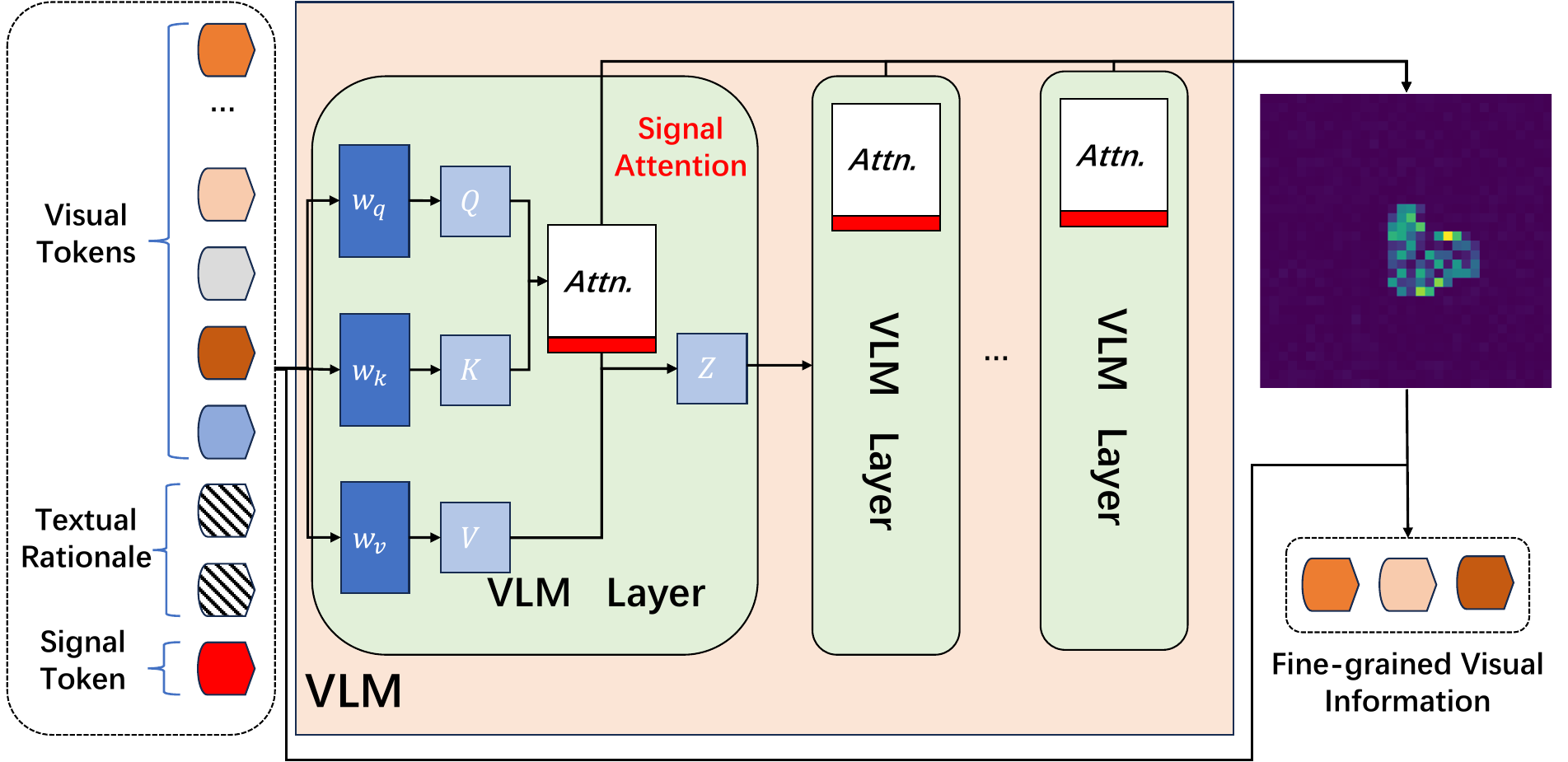}
    \caption{The workflow of ADS selecting fine-grained visual rationales. Signal attention represents the attention map of the signal token overall visual tokens.}
    \label{fig:ads}
\end{figure}

\subsection{Attention-driven Selection}
\label{sec:ads}
Although the proposed ICoT is both novel and conceptually sound, current VLMs are unable to generate such fine-grained visual information.
This remains true even for VLMs~\cite{meta2024chameleon,lu2024unified,wang2024emu3} that are empowered with multimodal generation ability.
We thus propose to simplify the problem from fine-grained visual information generation to fine-grained visual information selection, as this information has been naturally contained in the origin image, namely $x_v^{h' \times w'} \in x_v^{h\times w}$ where $h' \ll h$ and $w' \ll w$.

Specifically, before performing next-token prediction, ICoT requires the VLM to cache visual tokens $f_v^{l\times d}$ extracted by its built-in visual encoder $\mathbf{E}(x_v)$ for further selection.
In the following decoding steps, we consider the VLM deems it necessary to insert a piece of visual rationales after generating a pre-defined signal token $\mathcal{S}$ as shown in Figure~\ref{fig:ads}, which is a natural language token that indicates the beginning of a textual rationale.
Therefore, ADS will be signaled to select fine-grained visual information from $f_v^{l\times d}$ upon the VLM generating $\mathcal{S}$:
\begin{equation}
    \mathrm{do\_selection} = 
    \begin{cases} 
      \text{True} & \text{if } \mathrm{predicted\_tokens[-1]} = \mathcal{S}; \\
      \text{False} & \text{otherwise}.
    \end{cases}
\end{equation}
Then, ADS selects $n$ visual tokens from $f_v^{l \times d}$ as fine-grained visual information according to the signal attention map $A_t$ at the current decoding step $t$:
\begin{equation}
    \mathcal{V}_{\mathrm{selected}} = \{f_v[i] | i \in \mathrm{TopK}(A_t, n)\},
\end{equation}
where $A_t$ is obtained by averaging the attention map between the signal token and visual tokens across all VLM layers.
Up to now, the selected visual tokens are sorted by their attention scores, and we subsequently restore the relative position of $f_v[i]$ in the origin image in place, prioritizing rows.
Once fine-grained visual tokens $\mathcal{V}_{\mathrm{selected}}$ are obtained, VLMs will take as input the concatednated $\mathcal{V}_{\mathrm{selected}}$ and $\mathrm{predicted\_tokens}$, denote as $\mathrm{Cat}(\mathrm{predicted\_tokens}, \mathcal{V}_{\mathrm{selected}})$, resuming the original autoregressive text generation process.
Hence, the current decoding step is formulated as:
\begin{equation}
\begin{aligned}
    \mathrm{next\_token} = \mathbf{VLM}(\mathrm{Cat}(\mathrm{predicted\_tokens}, \\ \mathcal{V}_{\mathrm{selected}})).    
\end{aligned}
\end{equation}
Notably, to avoid misunderstanding, we first convert $\mathrm{predicted\_tokens}$ into word embeddings $f_e$, and then concatenate $f_e$ with selected fine-grained visual information $\mathcal{V}_{\mathrm{selected}}$ in the embedding-level.
We provide a detailed description of ICoT and ADS in Algorithm~\ref{alg:icot} and Algorithm~\ref{alg:ads}, respectively.

Technically, ICoT inherits the existing eliciting methods from CoT, such as attaching an instruction: ``Let's think step by step'' in zero-shot ICoT or providing few-shot examples.
In the few-shot ICoT, each example consists of an image, textual rationales, and visual rationales.
These examples can be manually designed to prompt VLMs regarding their way of thinking and generation formatting, among other aspects.

%% file: sec/4_experiments.tex
\section{Experiments}
\subsection{Datasets}
\textbf{M$^3$CoT}~\cite{chen2024m} is a novel multimodal CoT benchmark specifically concentrated on multi-domain, multi-reasoning-step.
M$^3$CoT contains 267 categories from science, mathematics, and commonsense domains.
As the question of each instance is relatively complex, their rationales have an average length of 293 tokens and rely more on fine-grained visual information, which can reflect the advantages of ICoT compared with previous multimodal CoT methods.

\textbf{ScienceQA}~\cite{saikh2022scienceqa} is a popular dataset used to evaluate the reasoning ability of VLMs. We use ScienceQA to provide a general comparison between ICoT and other existing multimodal CoT methods.

\textbf{LLaVA-Bench In-the-Wild} (LLaVA-W)~\cite{liu2024llavanext} evaluates VLMs' ability to respond to visual questions with detailed long-form answers, which also focus on the fine-grained visual description. The reference label of each instance is produced by GPT-4v.

\subsection{Baselines}
\textbf{No-CoT} responds to the current input image and question directly without further prompting.
The few-shot demonstrations of the direct generation mode consist of (Image, Question, Answer).

\textbf{Multimodal CoT}~\cite{zhangmultimodal} elicits VLMs to generate a series of text-only intermediate reasoning steps to infer the final outputs.

\textbf{CCoT}~\cite{mitra2024compositional} first generates a scene graph (SG) using the VLM itself and then uses that SG in the prompt to produce a response.
The SG is a JSON-like structural description of the given image with extensive compositional information of objects in the current images.
Following their settings, we apply their official prompt to prompt VLMs to generate SGs and answers respectively.

\textbf{DDCoT}~\cite{zheng2023ddcot} first prompts LLM to deconstruct the input question into a sequence of basic sub-questions and then applies a VQA model to answer these sub-questions involving visual information.
In this paper, we use the VLM that plays the role of LLM in DDCoT for a fair comparison, as their original LLM is ChatGPT.

\textbf{SCAFFOLD}~\cite{lei2024scaffolding} overlays a coordinate matrix onto the input image, exactly demonstrating relative visual positions for VLMs.
During reasoning, VLMs are steered to utilize these coordinates that indicate fine-grained visual information in the image to solve different vision-language tasks.
We use their released scripts to add coordinates over each image and then use their official prompt to elicit VLMs.

Specifically in the few-shot scenario, the demonstrations of these baselines are human-written, aligning with ICoT.

\begin{table*}
\begin{tabular}{l|l|ccc|ccc}
\hline
\multirow{3}{*}{Backbone} & \multirow{3}{*}{Methods} & \multicolumn{3}{c|}{\textit{0-shot}} & \multicolumn{3}{c}{\textit{1-shot}} \\
\cline{3-8}      &       & M$^3$CoT & ScienceQA & \multicolumn{1}{l|}{LLaVA-W} & M$^3$CoT & ScienceQA & \multicolumn{1}{l}{LLaVA-W} \\
      &       & ACC. $\uparrow$  & ACC. $\uparrow$  & ROUGE-L$\uparrow$ & ACC. $\uparrow$  & ACC. $\uparrow$  & ROUGE-L$\uparrow$ \\
\hline
\multirow{7}{*}{\textit{\textbf{Chameleon-7B}}} & No-CoT & 29.1  & 47.7  & 13.1  & 28.4  & 48.5  & 23.9 \\
& Multimodal CoT~\cite{zhangmultimodal}   & 28.5  & 49.0  & 20.4  & 30.6  & 50.7  & 20.6 \\
& CCoT~\cite{mitra2024compositional}  & 29.4  & 50.2  & 22.1  & 31.4  & 51.3  & 24.5 \\
& DDCoT~\cite{zheng2023ddcot} & 28.6  & 49.8  & 20.2  & 29.8  & 49.2  & 23.1 \\
& SCAFFOLD~\cite{lei2024scaffolding} & 29.6  & 48.5  & 21.7  & 31.1  & 47.5  & 24.7 \\
& ICoT (Ours)  & \textbf{29.8} & \textbf{51.0} & \textbf{25.2} & \textbf{32.3} & \textbf{53.4} & \textbf{27.6} \\
& \% \textbf{Improve} & 0.6\% &  1.6\% & 14.0\%  & 2.8\%  & 4.0 \%  & 11.7\% \\
\hline
\hline
\multirow{3}{*}{Backbone} & \multirow{3}{*}{Methods} & \multicolumn{3}{c|}{\textit{0-shot					}} & \multicolumn{3}{c}{\textit{1-shot}} \\
\cline{3-8}      &       & M$^3$CoT & ScienceQA & \multicolumn{1}{l|}{LLaVA-W} & M$^3$CoT & ScienceQA & \multicolumn{1}{l}{LLaVA-W} \\
      &       & ACC. $\uparrow$  & ACC. $\uparrow$  & ROUGE-L$\uparrow$ & ACC. $\uparrow$  & ACC. $\uparrow$  & ROUGE-L$\uparrow$ \\
\hline
\multirow{7}{*}{\textit{\textbf{Qwen2-VL-7B}}} & No-CoT & 43.6  & 56.3  & 32.7  & 45.4  & 64.4  & 33.5 \\
  & Multimodal CoT~\cite{zhangmultimodal}   & 40.1  & 51.3  & 30.7  & 42.5  & 58.3  & 31.4 \\
  & CCoT~\cite{mitra2024compositional}  & 43.3  & 56.4  & 29.4  & 44.1  & 63.8  & 33.9 \\
  & DDCoT~\cite{zheng2023ddcot} & 42.6  & 55.2  & 31.2  & 45.7  & 64.9  & 32.8 \\
  & SCAFFOLD~\cite{lei2024scaffolding} & 41.7  & 53.7  & 31.8  & 44.9  & 62.5  & 33.1 \\
  & ICoT (Ours) & \textbf{44.1} & \textbf{56.8} & \textbf{34.2} & \textbf{46.0} & \textbf{65.4} & \textbf{35.7} \\
  & \% \textbf{Improve}  & 1.1\%  & 0.7\%  & 4.6\%  & 0.6\%  & 0.7\%  & 5.3\% \\
\hline
\end{tabular}%
\caption{Results of ICoT and baselines based on Chameleon and Qwen2-VL, with the highest score \textbf{bold}.
M$^3$CoT and ScienceQA are evaluated by accuracy, and we report the ROUGE-L score for the LLaVA-W benchmark.
\% \textbf{improve} represents the relative improvement achieved by ICoT over the previously best baseline.}
  \label{tab:mainres}%
\end{table*}

\subsection{Implement Details}
We apply ICoT over Chameleon-7B~\cite{meta2024chameleon} and Qwen2-VL-7B-Instruct~\cite{Qwen2VL}, which represents the fine-grained visual information in the form of discrete vokens and dense features.
All experiments are conducted on A800 GPUs, and we evaluate ICoT under both zero- and one-shot scenarios.
During generating interleaved-modal rationales, the signal token $\mathcal{S}$ used to trigger ADS is set to line break, i.e., ``\textbackslash n'', by default, which semantically and empirically indicates the end of a generated rationale and the beginning of the next one.

VLMs insert visual tokens selected by ADS following the special token at the granularity of 64 according to posterior results shown in Table~\ref{fig:number} of Appendix~\ref{sec:hyper}.
Notably, to shorten the representation of an image, Qwen2-VL introduces a novel merge mechanism, and we approximately consider its patch size to be $(28 \times 28)$.
Each patch of Qwen2-VL has approximately 4 times as many pixels as a Chameleon patch $(16 \times 16)$, which results in practical selection numbers of ADS set to 16.
Considering the work of ADS requires the inner attention map, we apply the ``eager'' attention on both Chameleon-7B and Qwen2-VL, limited to the dependency of related python libraries~\footnote{Using attn\_implementation=``eager'' when loading the model from HuggingFace.}.

\subsection{Main Results}
We comprehensively evaluate the performance of ICoT on top of Chameleon-7B and Qwen2-VL-7B through M$^3$CoT, ScinceQA, and LLaVA-W in Table~\ref{tab:mainres}.
In 0-shot settings, ICoT outperforms all baselines, including direct generation (No-CoT), CoT, CCoT, DDoT, and SCAFFOLD.
Specifically, ICoT distinguishes from Multimodal CoT in terms of the modality of reasoning steps, which exhibit the advantages of interleaved-modal rationales to infer the final answer effectively.
Compared with other multimodal CoT methods, the performance gains of ICoT further indicate that interleaved-modal rationales are more reasonable in intuition and effect than plainly inserted scene graphs (CCoT) and deconstructed sub-questions (DDoT).
In 1-shot settings, ICoT demonstrations contain manually selected fine-grained visual information, while their text rationales are kept the same as other baselines.
The performance gains compared with 0-shot ICoT indicate that our manually designed fine-grained ICoT demonstrations potentially guide VLMs to think in this format.
In Table~\ref{tab:demo}, we rigorously ablate the effectiveness of fine-grained ICoT demonstrations.

Additionally, ICoT achieves the most relative performance gains in the LLaVA-W benchmark as the reference labels contain details sourced from images.
These substantial performance gains compared with other baselines prove that visual tokens selected by ADS effectively capture the fine-grained visual information of an image, aiding the generation of high-quality text rationales.

\subsection{Ablation Study}
We ablate ICoT to verify the effectiveness of each portion across three benchmarks in Table~\ref{tab:io} with the following settings: (1). w/o ADS: VLMs generate text-only rationales. (2). w/o FVI: Patches inserted in the demonstration are randomly sampled.
Results indicate that both ADS and fine-grained visual information (FVI) incorporated in the demonstration are necessary.
In particular, interleaved-modal rationales exhibit substantial advantages in generating high-quality textual rationales compared with text-only rationales (w/o ADS).
When substitute ICoT demonstration with normal ones (w/o FVI), the performance degradation proves the fact that fine-grained visual information in demonstrations effectively guides VLMs to think in this format.
Compared with the performance difference between removing ADS and FVI, we find that generating paired visual and text rationales boosts more improvements.

Additionally, the performance gap is relatively smooth in ScienceQA and more dramatic on M$^3$CoT and LLaVA-W.
We attribute this to the ScienceQA dataset being relatively easier than others since both M$^3$CoT and the answers of LLaVA-W highly rely on the fine-grained visual information of an image.
Therefore, our proposed ICoT has the potential to solve complex vision-language tasks.

\subsection{In-depth Analysis}
\paragraph{Analysis on realizing ICoT via KV Cache}
Up to now, the fine-grained visual information of ICoT is provided at the input end via discrete vokens or dense visual tokens, which brings more computation.
After rethinking the generating process of an autoregressive model, there are other inputs that are proposed to avoid repeated computation, namely, the Key-Value (KV) Cache.
The input image was stored in the KV Cache during the prefilling phase before generating multimodal intermediate reasoning steps in a left-to-right fashion.
Therefore, copying the KV cache of fine-grained visual information enables ICoT with reduced computational costs, as visual information does not require extra forward propagation.
As shown in Table~\ref{tab:kv}, copying the KV Cache brings performance degradation compared with providing visual information at the input end.
We attribute this phenomenon to the fact that although copying KV Cache indeed makes VLMs attend more to the same region as ADS, the optimal visual information is highlighted in a \textbf{position-agnostic} case, determined by the nature of KV Cache.
Specifically, this degrades the original interleaved-modal rationales into non-interleaved ones as position information is early fused into KV Cache, and thus the copied ones are inherently insensitive to the position of textual rationale.

\begin{table}[t!]
  \centering
  \resizebox{0.98\linewidth}{!}{
    \begin{tabular}{l|ccc}
    \bottomrule
    Methods & M$^3$CoT & ScienceQA & LLaVA-W \\
    \hline
    ICoT & 32.3  & 53.4  & 27.6 \\
     w/o ADS & 29.2 (\textcolor{red}{-3.1}) & 52.4(\textcolor{red}{-1.0}) & 24.5(\textcolor{red}{-3.1}) \\
     w/o FVI & 30.6 (\textcolor{red}{-1.8}) & 52.8(\textcolor{red}{-0.6}) & 25.9(\textcolor{red}{-1.7}) \\
     w/o ADS+FVI & 29.1(\textcolor{red}{-3.2}) & 51.0(\textcolor{red}{-2.4}) & 23.0(\textcolor{red}{-4.6}) \\
    \toprule
    \end{tabular}}
  \caption{
  Ablation studies of \textit{1-shot} ICoT on Chameleon-7B.
  () describes the performance degradation compared with ICoT.
  FVI indicates the 1-shot demonstration contains fine-grained visual information.
  ADS indicates that VLMs generate interleaved-modal reasoning steps.
  }
  \label{tab:io}%
\end{table}%
\begin{table}[t!]
  \centering
  \resizebox{0.98\linewidth}{!}{
    \begin{tabular}{l|cc|cc}
    \bottomrule
    \multicolumn{1}{c|}{\multirow{2}[4]{*}{Dataset}} & \multicolumn{2}{c|}{\textit{0-shot}} & \multicolumn{2}{c}{\textit{1-shot}} \\
\cline{2-5}          & KV-Copy & ICoT  & KV-Copy & ICoT \\
    \hline
    M$^3$CoT & 29.1  & 29.8  & 31.5 & 32.3 \\
    ScienceQA & 49.7  & 51.0  & 52.9  & 53.4 \\
    LLaVA-W & 24.7  & 25.2  & 27.0  & 27.6 \\
    \toprule
    \end{tabular}}
  \caption{Results comparison between copying KV Cache (KV-Copy) and inserting selected patches.}
  \label{tab:kv}%
\end{table}%

However, considering it brought slight performance degradation and factually reduced computation costs, we believe this exploration is still valuable, and we call for more interesting exploration in realizing ICoT.

\paragraph{Analysis on the Demonstrations}
\begin{table}[t!]
    \centering
    \resizebox{0.98\linewidth}{!}{
    \begin{tabular}{l|ccc}
    \bottomrule
    Methods & M$^3$CoT & ScienceQA & LLaVA-W \\
    \hline
    Human-written & 32.3  & 53.4  & 27.6 \\
    Model-written  & 31.5(\textcolor{red}{-0.8}) & 51.8(\textcolor{red}{-1.6}) & 26.7(\textcolor{red}{-0.8}) \\
    \toprule
    \end{tabular}}
    \caption{
    Results concerning the design of demonstrations on Chameleon-7B.
    () describes the performance degradation compared with ICoT.
    Human-written indicates that demonstrations are manually designed with fine-grained visual information inserted, and Model-written indicates that the VLM generates both visual and textual rationale via ICoT.}
  \label{tab:demo}%
\end{table}%
\begin{figure*}[t!]
    \centering
    \includegraphics[width=0.98\linewidth]{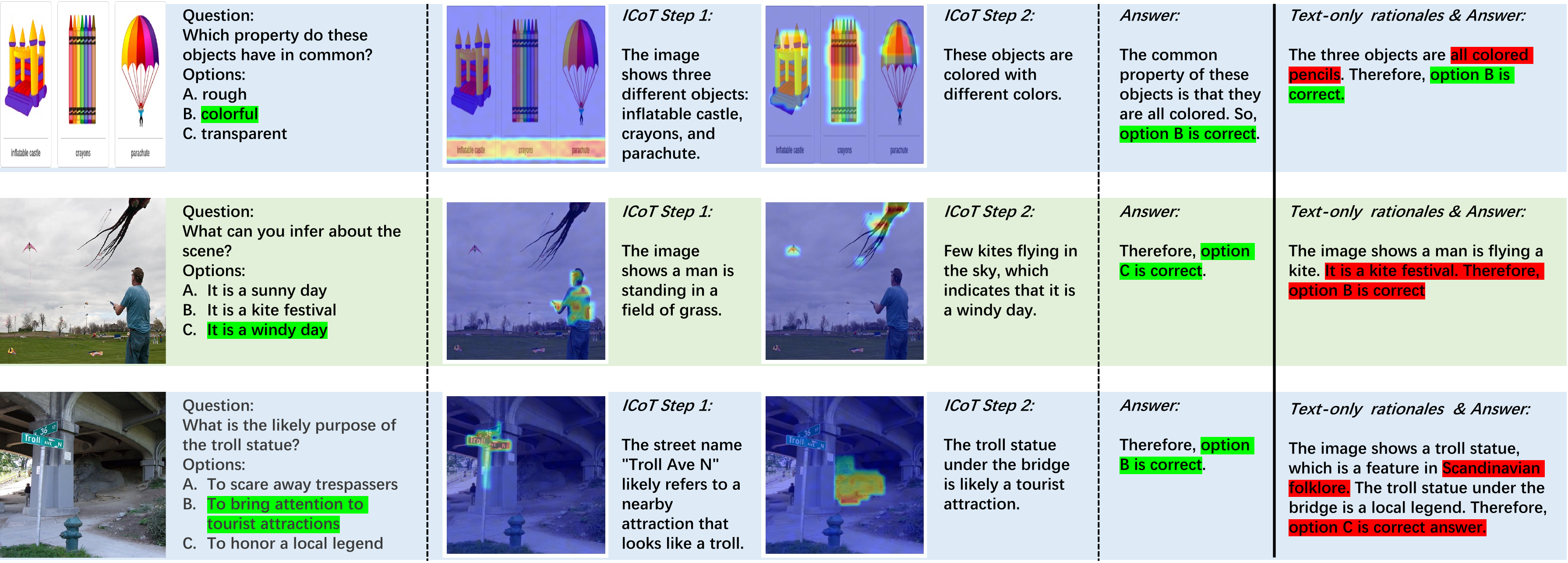}
    \caption{Case studies between ICoT and multimodal CoT with text-only rationales on Chameleon. Three cases are selected according to three typical problems in text-only problems: misunderstanding, overgeneralization, and hallucination. \textcolor{red}{Red} blocks indicates the incorrect rationales.}
    \label{fig:casestudy}
\end{figure*}

We also attempt to let VLMs generate the demonstrations via themselves (Automatic at the bottom of Table~\ref{tab:demo}).
Results indicate that using the automatically generated demonstrations also brings performance degradation compared with ICoT using manually designed ones.
We consider it is caused by the fact that formulating a continuous sub-image through ADS is non-trivial, and some discrete patches inevitably introduce additional noise.
Therefore, considering that designing such a demonstration is not time-consuming, ICoT utilizes manually designed ones to elicit VLMs to perform ICoT for better performance.

%% file: sec/6_casestudy.tex
\section{Case Study}

In this section, we empirically illustrate the advantages of ICoT via three case studies in Figure~\ref{fig:casestudy}.
These case studies focused on three typical problems that occurred in text-only rationales, namely, misunderstanding (top), overgeneralization (middle), and hallucination (bottom).

Specifically, in the first case, interleaved-modal CoT first recognizes three different objects via captions: ``inflatable castle,
crayons, and a parachute''.
Then, in the second reasoning step, ADS inserts selected patches from the scheduled objects to elicit the VLM to conclude their common property, and VLM infers a correct answer.
Text-only CoT misunderstands the three objects are all colored pencils, ignoring the castle and the parachute, even the final answer is correct.
In the second case, text-only rationales overgeneralize flying a kite to a kite festival, leading to a wrong answer.
ICoT first recognizes a man standing in a field of grass and then infers it is a windy day according to a few kites in the sky.
In the last case, it provides the other typical error of text-only CoT, namely, hallucination.
As text-only CoT purely relies on language reasoning ability, VLMs have the potential to imagine something not mentioned in the image, resulting in a wrong answer.
ICoT first infers from the street sign that there may be a troll attraction nearby according to patches of the indicator inserted by ADS. 
Then, the ADS helps the VLMs to attend to the troll statue under the bridge and infer it is likely the mentioned attraction, finally arriving at the correct answer.

Even though the above case studies exhibit the advantages of ICoT, ADS still brings potential problems.
For example, ADS is triggered to select patches when VLM generates a pre-defined signal token.
This simple mechanism is a double-edged sword that VLMs will generate low-quality responses if this token is generated with a high frequency.

%% file: sec/7_conclusion.tex
\section{Conclusion}
In this paper, we first propose interleaved-modal CoT (ICoT), which generates interleaved-modal rationales to infer the final answer effectively.
In light of the challenges of applying ICoT on existing VLMs, we then introduce Attention-driven Selection (ADS), a plug-and-play strategy to identify optimal patches from the image without being parameterized.
We evaluate ICoT on Chameleon-7B and Qwen2-VL-7B-Instruct, representing VLMs of two architectures.
Extensive experiments conducted on M$^3$CoT, ScienceQA, and LLaVA-W, under both zero- and few-shot scenarios, have proven that ICoT achieves substantial performance (up to 14\%) compared with the existing multimodal CoT methods.
Additionally, in the analysis section, we conduct a preliminary exploration of implementing ICoT by copying the KV cache of optimal visual tokens and explain the inner trade-off between efficiency and performance in this approach.

Although ICoT has proven its effectiveness in this paper, we consider ICoT still has significant potential for further improvement.
The patch selection in ADS requires storing attention scores, which brings additional memory overhead.
Moreover, the fixed number of selected patches in the ADS design is sub-optimal, resulting in unexpected outputs for VLMs.
To address these, we intend to incorporate established techniques from segmentation or grounding methods to create a more robust implementation of ICoT.
In the future, we also plan to evaluate it across additional backbones and benchmarks to better assess its generalization ability.

\section{Acknowledgement}
I would like to express my sincere gratitude to all the authors and reviewers for their valuable contributions to this research.
This work was supported by the National Natural Science Foundation of China (NSFC 62106165) and the Project Funded by the Priority Academic Program Development of Jiangsu Higher Education Institutions, China.

%% file: sec/X_suppl.tex
\clearpage
\setcounter{page}{1}
\maketitlesupplementary

\section{Analysis on the Selected Patches}
\label{sec:hyper}
\begin{figure}[!t]
    \centering
    \includegraphics[width=0.99\linewidth]{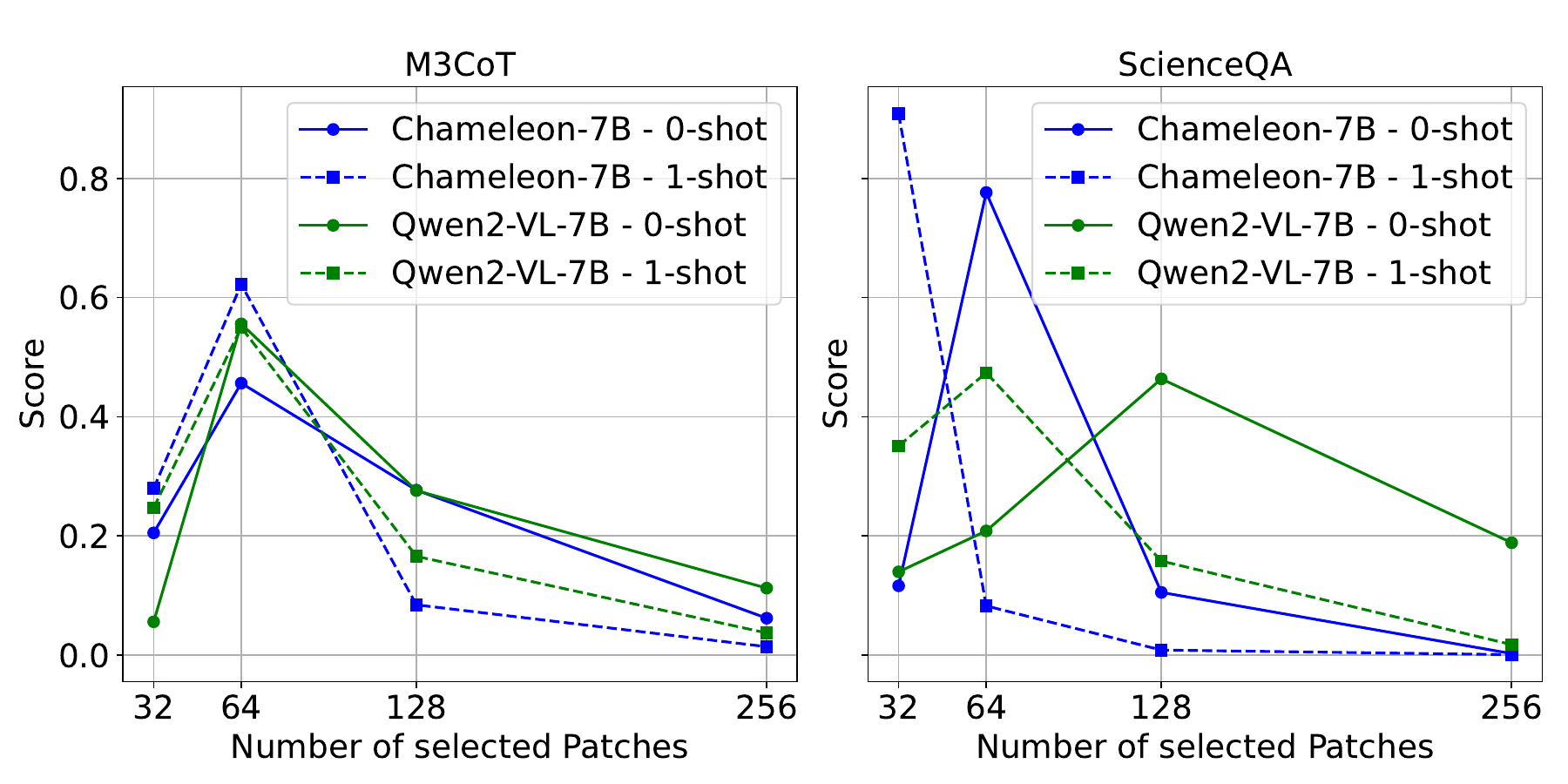}
    \caption{The results of ICoT across validation sets of two datasets on both Chameleon and Qwen2-VL, with the number of selected patches set to 32, 64, 128, and 256. The reported scores are normalized for simplicity.}
    \label{fig:number}
\end{figure}

Intuitively, the performance of ICoT is sensitive to the number of selected patches. 
If ADS selects a large number of patches every time, the selected patches will be dispersed, resulting in more noise introduced and higher computation costs.
In contrast, only a few selected patches perhaps failed to contain enough fine-grained visual information.
It is non-trivial to determine the exact number of patches selected by ADS, as fine-grained information in an image is not always the same size.
Therefore, in Figure~\ref{fig:number}, we empirically set the number of patches selected by ADS $n$ to 32, 64, 128, and 256 at a coarse-grained level and illustrate their performance variance across two benchmarks~\footnote{LLaVA-W only contains a test set. }.
Observed results indicate that setting $n$ too large or too small is not good for VLMs, and ICoT achieves relatively better performance when $n$ is set to 64.

\section{Performance on General Benchmark}
\begin{table}[h]
    \centering
    
    \resizebox{0.98\linewidth}{!}{
    \begin{tabular}{l|ccc}
    \bottomrule
    1-shot & Flickr30k (CIDEr $\uparrow$)& OKVQA (VQA-ACC 
$\uparrow$) \\
    \hline
    Chameleon & 22.3  & 26.2  \\
    \hline
    \quad +ICoT  & 23.6 & 28.2   \\
    \toprule
    \end{tabular}}

    \caption{Evaluation on general benchmarks}

  \label{tab:general}%
\end{table}%
ICoT is a plug-and-play prompting method designed for complex multimodal reasoning, while the performance of ICoT on tasks requiring weak reasoning ability is still unknown.
To explore whether ICoT causes degradation, we evaluate ICoT on captioning and VQA in Tab.\ref{tab:general}.
Results indicate advantages of ICoT.

\begin{figure}[!t]
    \centering
    \includegraphics[width=0.99\linewidth]{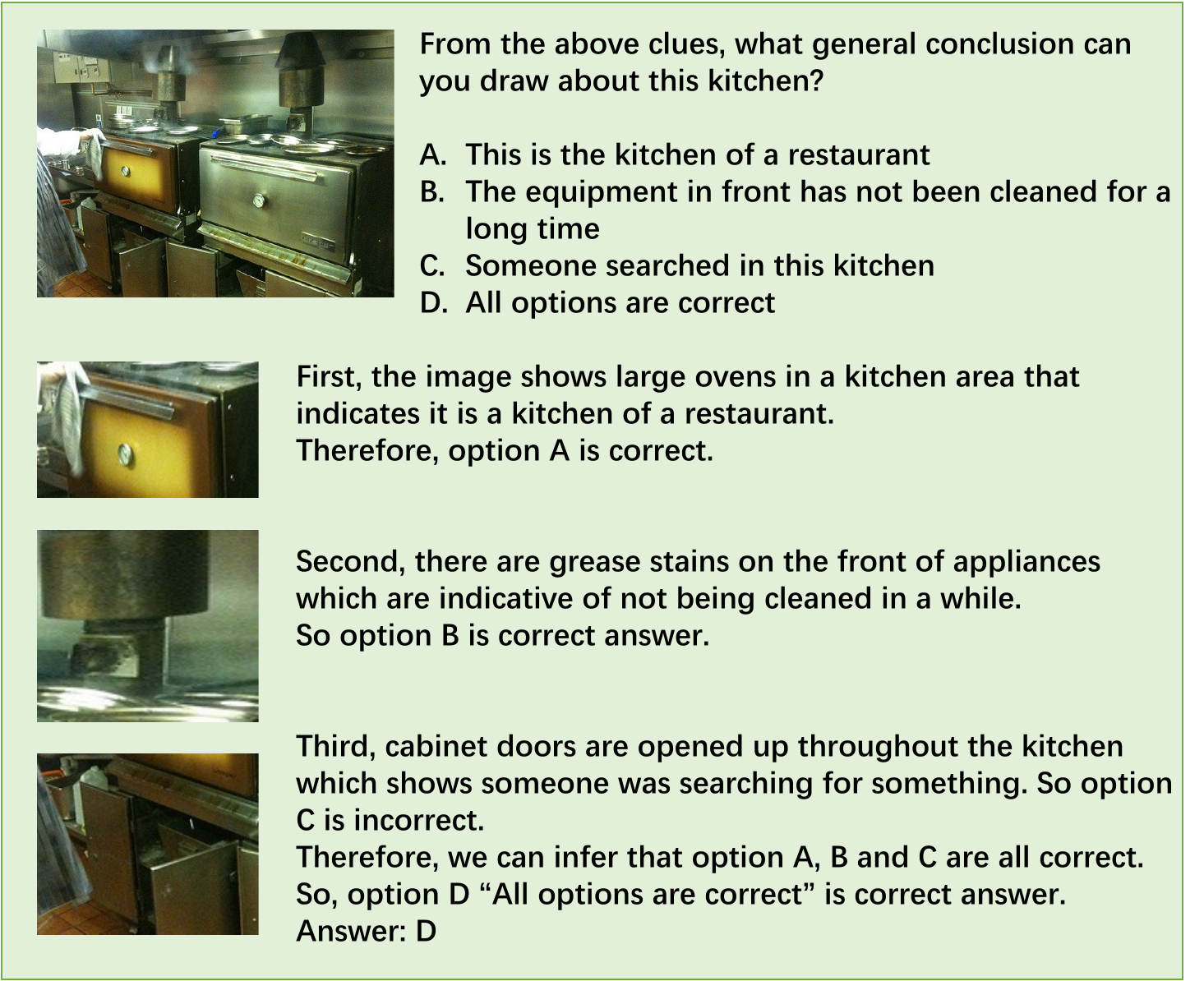}
    \caption{The case of demonstration with Fine-grained Visual Information (FVI), which is used in 1-shot ICoT. }
    \label{fig:fvi_case}
\end{figure}

\section{Detail Declaration}
In Fig.~\ref{fig:fvi_case}, we provide a case to illustrate the FVI in 1-shot ICoT.
In Algorithm 1, the stopping criteria is maximum generation length or generating the special token of ``end of sequence''.